\documentclass{bmvc2k}
\usepackage{booktabs}
\usepackage{hyperref}
\usepackage{amsmath}
\usepackage{wrapfig}
\usepackage{amsfonts}

\title{BaseTransformers: Attention over base data-points for One Shot Learning}

\addauthor{Mayug Maniparambil}{mayugmaniparambil@gmail.com}{1,2}
\addauthor{Kevin McGuinness}{kevin.mcguinness@dcu.ie}{1,2}
\addauthor{Noel O'Connor}{noel.oconnor@dcu.ie}{1,2}

\addinstitution{
 ML Labs\\
 SFI Centre for Research Training,
 Dublin, Ireland
}
\addinstitution{
 Dublin City University,
 Dublin, Ireland
}

\runninghead{Maniparambil, McGuinness, O'Connor}{BaseTransformers}

\begin{document}

\maketitle

\begin{abstract}
Few shot classification aims to learn to recognize novel categories
using only limited samples per category. Most current few shot methods
use a base dataset rich in labeled examples to train an encoder that
is used for obtaining representations of support instances for novel classes. Since
the test instances are from a distribution different to the base
distribution, their feature representations are of poor quality,
degrading  performance. In this paper we propose to make
use of the well-trained feature representations of the base dataset that
are closest to each support instance to improve its representation
during meta-test time. To this end, we propose BaseTransformers, that attends to the most relevant regions of the base dataset
feature space and improves support instance representations. Experiments on three benchmark data
sets show that our method works well for several backbones and
achieves state-of-the-art results in the inductive one shot setting. Code is available at  \href{https://github.com/mayug/BaseTransformers}{github.com/mayug/BaseTransformers} .
\end{abstract}

\section{Introduction}
\label{sec:intro}
 The development of few shot learning models is important for real world deployment of artificial vision systems outside of controlled scenarios. Most previous works focus on developing stronger models, while scant attention has been paid to the properties of the data itself and the fact that as the number of data points increase, the ground truth distribution can be better uncovered. Estimating the prototype for a novel class using a single instance is fundamentally ill posed, resulting in poor one shot performance. \cite{yang2021free} has shown that this can be alleviated by modeling the class conditional distribution as a Gaussian and sampling a large number of features from this distribution to train a classifier or estimate the prototype. They show that distributions of semantically similar classes in the base dataset have similar mean and variance to the distributions of the novel class. Therefore, the statistics of the class conditional distributions of novel classes are transferred from those of base classes which have been estimated with several examples (over 600) per class. This method assumes that the class conditional feature distributions are uni-modal Gaussian and that the transferable statistics are only global and not local to each base instance or its spatial locations. 

We propose a novel method for estimating prototypes of unseen classes using the base dataset without making any assumptions on the distribution of the base data feature space or the transferability of the instance level or spatial level information. Our proposed method, BaseTransformers, is an end-to-end learnable cross attention mechanism that estimates a robust, base aligned prototype for novel categories by learning local part based correspondences between the support instance and semantically similar base instances. This is based on two key ideas: (i) the base dataset images are composed of semantically meaningful parts that could be reused during the classification of novel images; and (ii) since the base data features are estimated using many shots, the features corresponding to these parts are less noisy representations, closer to the ground truth distribution. The concept is illustrated in  Fig 1, where a novel `centaur' class  has an undersupported prototype in the feature space of an encoder pretrained on base-data. However a robust prototype of a centaur can be constructed by taking the head, torso of a human and the body and legs of horse base classes which are individually well supported in the feature space. 

We hypothesize that semantically similar parts of a well represented base data feature space can be used to estimate a novel prototype that is effectively a part based composition of the well estimated base data regions. To enable this BaseTransformers allow for: (i) spatial part based comparison between the support instance and similar base instances to select the semantically meaningful regions of the robust base data feature space; and (ii) aggregation of the semantically similar parts of the base instances to estimate a novel prototype that is a composition of robust meaningful base regions. Taking inspiration from \cite{doersch2020crosstransformers} we instantiate a cross attention mechanism on the feature space of the pretrained encoder to enable this. We perform this adaptation of the support instance using the base instances in the feature space and not the original pixel space, as the feature space has lower dimensions and  semantically meaningful structures that are more easily transferable between the base and novel domains.

\begin{figure}
\centering
\includegraphics[width=10cm]{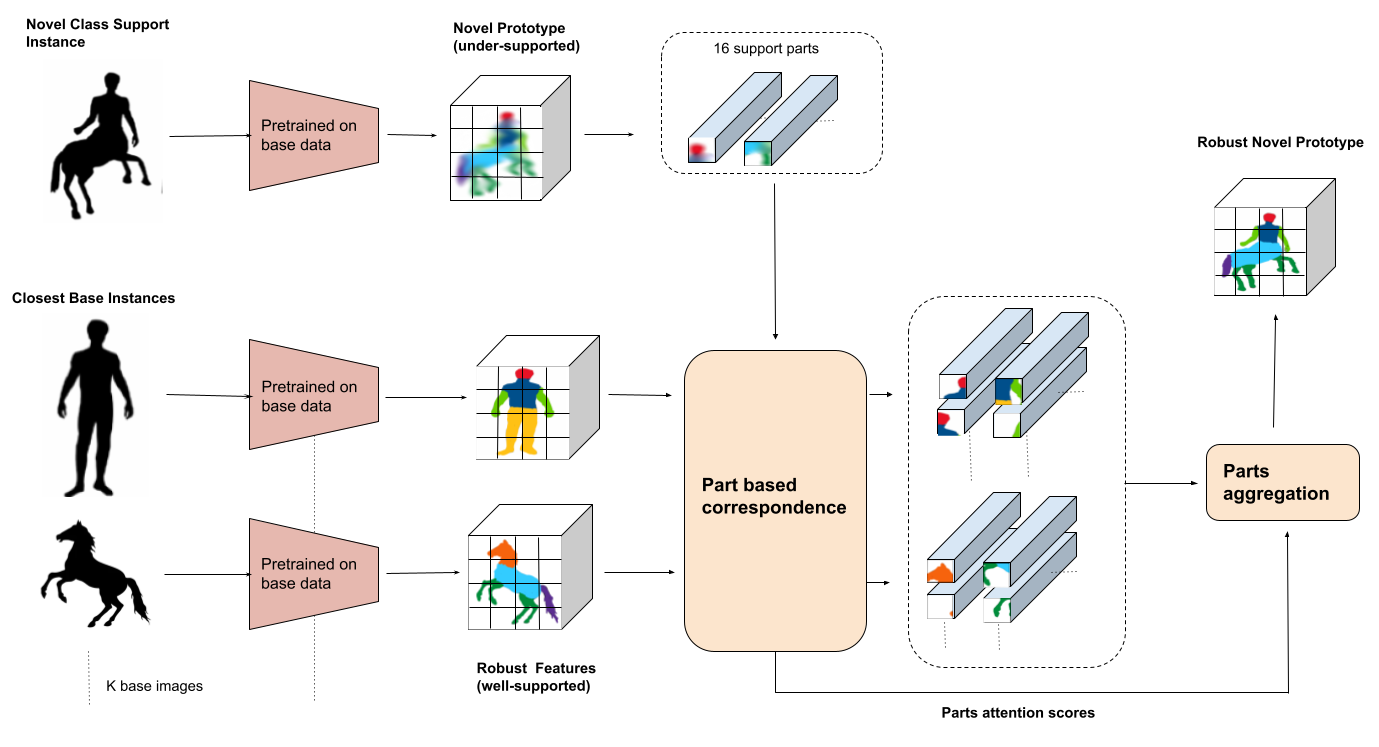}
\caption{BaseTransformers construct robust novel class prototypes by attending to and aggregating semantically similar regions of the well supported base data feature space instead of using the noisy novel prototype as in Prototypical Networks \cite{snell2017prototypical}.}
\label{fig:illustrative}
\end{figure}

For each episode, the BaseTransformer takes the 2D feature spaces of the support instance as query and the closest base instances as the key and value. The BaseTransformer is trained end-to-end using the meta learning paradigm to identify the most relevant regions in the base data feature space and use them to compose a more robust novel class prototype. Our approach starts with a pretraining stage using cross entropy and contrastive losses on the base dataset to produce a robust encoder bypassing supervision collapse \cite{doersch2020crosstransformers, mangla2020charting}. This is followed by a meta training stage, in which the encoder and BaseTransformer are jointly trained to adapt the support instances using instances from the closest base classes. To identify the closest base classes we propose using the class label information of the support instances, and making queries on the base dataset based on semantic similarity. We show that the proposed method beats the current state-of-the-art in 3 different datasets (70.88\%, 72.46\%, 82.27\% on mini-ImageNet, tiered-ImageNet and CUB respectively) in the inductive one shot setting.

Our novel contributions are: i) We identify that robust novel prototypes in one shot learning can be obtained by part based composition of semantically similar base features; ii) We design BaseTransformer that improves the 1-shot prototypes by learning to attend to the robust 2D feature space of base instances and aggregate these to compose the novel prototype; iii) We evaluate our method on two backbones and three benchmarks to show its effectiveness in the one shot inductive setting of few shot learning.

\section{Related Work}
\textit{Meta Learning} aims to extract common useful knowledge for classifying novel classes by emulating few shot tasks during training time, and are usually optimization based or metric learning based. In optimization based methods, the objective is to meta-learn a good initialization of weights \cite{finn2017model,rusu2018metalearning,rajeswaran2019meta,zintgraf2019fast} or the optimization process \cite{ravi2017optimization,li2017meta,munkhdalai2017meta,xu2020metafun} or a combination of both \cite{park2019meta,baik2020meta}. In metric learning methods \cite{vinyals2016matching,snell2017prototypical,sung2018learning,yoon2019tapnet} the objective is to develop an embedding space where similar instances are close to each other in some distance sense so that a simple nearest neighbour classifier can be used during meta test time. Our method is similar to metric learning, specifically prototypical networks, as we only have an extra transformer stage to adapt the support instances to form more robust prototypes. 

\textit{Transfer Learning} methods train a network to classify base classes, followed by finetuning the classifier on the novel instances whilst keeping the encoder fixed. ~\cite{wang2019simpleshot,chen2018a} has shown that this simple strategy performs surprisingly well, beating/matching several complex meta learning algorithms. We follow works such as~\cite{ye2020few} and have a pretraining stage in which the encoder is trained on a combination of cross entropy and self supervised loss. Other works~\cite{su2020does,gidaris2019boosting,mangla2020charting,liu2021learning} have shown that addition of self supervision losses in the pretraining stage provides more robust features, resulting in improved few shot performance. We use the InfoNCE loss \cite{chen2020simple} as an auxiliary loss during the pretraining stage.

A \textit{Base Dataset} has been used explicitly during meta test time in previous works, such as in ~\cite{yang2021free,afrasiyabi2020associative}. The approach of~\cite{yang2021free} models the feature space of each class as a Gaussian and transfers statistics from well estimated base class distributions to novel class distributions, and sample from this to train a classifier. In our approach, we do not assume that the class feature space follows a Gaussian distribution, but use a parametric function- a transformer to improve the prototype representation by means of attention over the feature space of base examples. The approach reported in~\cite{afrasiyabi2020associative} aligns the feature space of the novel instances to that of the closest base instances by reducing an adversarial alignment loss during the test time, while we do not tune any parameters of the transformer network during meta test time. Both methods make use of cosine similarity in the feature space to query the closest base classes. While this works well for us for shallow encoders, we find that making use of semantic information from the class labels results in semantically closer base classes.

\textit{Transformers} have also been investigated in a similar context. Previous works like \cite{ye2020few,hou2019cross} make use of transformer based adaptation on the feature space to improve few shot performance. The approach in \cite{ye2020few} uses self attention over the prototypes to adapt them in a task specific manner, while the approach of \cite{doersch2020crosstransformers} builds a classifier that aligns the prototypes and the queries spatially. Similarly \cite{kang2021relational, li2019revisiting, wu2019parn, hao2019collect} use different forms of self-correlation and cross-correlation mechanisms to improve the relational comparison between the prototypes and the query instances. We differ from these methods, in that we  explicitly attend over all spatial locations of a base data subset to improve the support instance features. To our knowledge, our work is the first to apply attention over the base data points for few shot learning.

\section{Method}
In this section we first introduce the setup of few shot classification in section \ref{sec:preliminaries} followed by description of our proposed method in sections \ref{sec:basetransformer} through \ref{sec:blind}.

\subsection{Preliminaries} \label{sec:preliminaries}
We follow the inductive setting for few shot learning. A few shot task is an $N$ way $M$ shot classification problem, with $N$ classes sampled from novel classes \( C_{n} \) with $M$ examples per class. \begin{math}D_\mathit{s} = \left \{x_i, y_i \right \}_{i=1}^{M\times N}\end{math} refers to the support set sampled from novel classes \( C_{n} \).
Test instances \(x_\mathit{q}\) are sampled from a a query set \(D_\mathit{q} = \left \{x_i, y_i \right \}_{i=1}^{Q}\) and the goal is to find a function $f$ that classifies  \begin{math}x_\mathit{q}\end{math} via \begin{math}\hat y = f\left ( x_{q}\mid D_\mathit{s} \right )\end{math}. In the few shot learning literature $M$ is usually 1 or 5 referring to the 1-shot or 5-shot task.

Finding $f$ from the very few examples in the support set is very difficult, so a base dataset is provided consisting of base classes \(C_b\) such that \begin{math}C_{b}\cap C_{n}=\emptyset \end{math}. In the meta-learning paradigm, $f$ is learnt by sampling several $N$-way $M$-shot tasks \(D_\mathit{s}^\mathit{b}\) and corresponding query sets \(D_\mathit{q}^\mathit{b}\) from the base dataset to emulate the test time scenario. In each sampled task, $f$ is learnt to  minimize the average error on \(D_\mathit{q}^\mathit{b}\):
%

\begin{equation}
f^{*}=\underset{f}{\arg\min} \sum_{(x_\mathit{q}^\mathit{b}, y_\mathit{q}^\mathit{b})\in D_\mathit{q}^\mathit{b}}^{} \ell\left(f(x_\mathit{q}^\mathit{b}\mid D_\mathit{s}^\mathit{b}),\, y_\mathit{q}^\mathit{b}\right),
\end{equation}

where $\ell$ can be any loss that measures the discrepancy between prediction and true label.

During meta test time the optimal \(f^{*}\) is applied on tasks sampled from \(C_{n}\). The performance of the model is evaluated on multiple tasks sampled from the novel classes \(C_{n}\).
For example, in prototypical networks, $f$ consists of an embedding network $E$ and a nearest neighbour classifier:
\begin{equation}
\phi_{x} = E(x) \in \mathbb{R}_{}^{d},
\qquad
\hat{y}_\mathit{q} = f(\phi_{x_\mathit{q}}; \{\phi_{x_\mathit{s}}^c\}),
\end{equation}
where \(\{\phi_{x_\mathit{s}}^c\}\) is the set of prototypes. Here, each prototype is given by: \begin{equation}\phi_{x_\mathit{s}}^c = \sum_{y_{i}\in c} E(x_{i}),\qquad \left (x_{i},y_{i}  \right )\in D_\mathit{s}.\end{equation}

\begin{figure}
\centering
\includegraphics[width=0.8\linewidth]{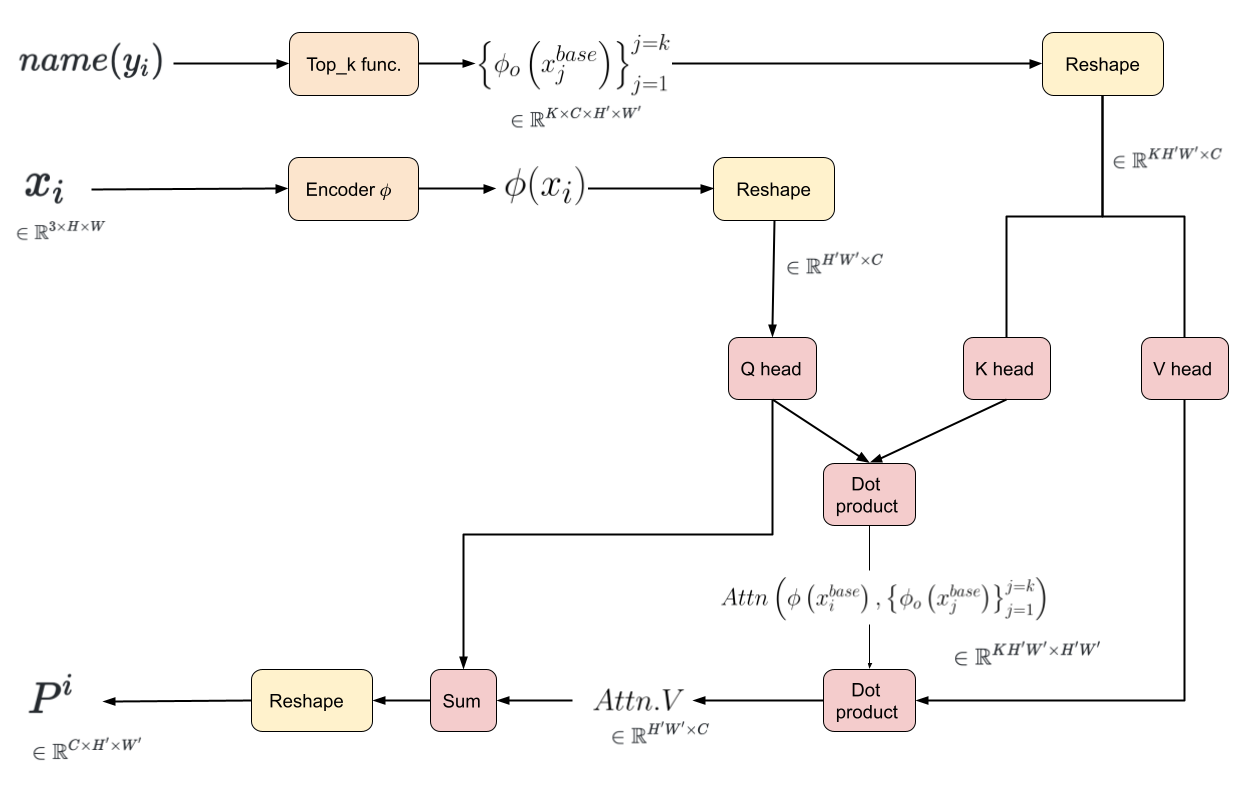}
\caption{Support instance feature \(\phi(x_{i})\) is reshaped and projected by query head Q to obtain queries \(q_{m}^{i}\) where m corresponds to spatial locations in the support instance. \(q_{m}^{i}\) is then compared with the keys \(k_{n}^{j}\) from all spatial locations \(n\) of base instances to get attention scores \(attn_{mjn}^{i}\), which are used to aggregate the values \(v_{n}^{j}\) and summed with original support feature \(\phi(x_{i})\) to obtain the base adapted prototype.}
\label{fig:arch_image}
\end{figure}

\subsection{BaseTransformer} \label{sec:basetransformer}
Given a support instance \(x_{i}\) and its closest base instances \( \left \{  x_{i}^\mathit{base}   \right \}_{i=1}^{k}
\) the BaseTransformer aims to learn a representation that enables part-based adaptation of  \(x_{i}\) by attending over all the spatial locations of all base instances in \( \left \{  x_{i}^\mathit{base}   \right \}_{i=1}^{k}.\)

First, an image representation of support instance \(\phi\left ( x_{i} \right )\) is obtained using the encoder \(\phi\), while the class name corresponding to the support instance is used to get the $k$ closest instances in the base dataset. The top $k$ function is described in detail in Section \ref{sec:queryingfunction}. The features of the closest base instances are passed through a fixed encoder \(\phi_{0}\) whose weights are the weights obtained after the pre-training stage on the base dataset. These representations are then used by the Transformer to establish correspondences between support instances and base instances to produce the adapted prototype. Finally, similar to prototypical networks, the Euclidean distance is used to classify the query feature \(\phi\left ( x_\mathit{test} \right )\) by making use of  adapted prototypes \(\left \{ P_{i}^\mathit{} \right \}_{i=1}^{N}\). Prototypical networks use 1D feature embedding while, BaseTransformers use 2D embeddings as input to allow the model to make part based soft correspondences between support and base instances, and use these to weigh the most relevant regions of base instances to estimate the prototype of a support instance as a composition of robust base parts.

More concretely, we consider a CNN without the final fully connected or pooling layers, such that \(\phi\left ( x_{i} \right )\in \mathbb{R}^{C\times H'\times W'}\). Top $k$ function uses the pre-trained encoder \(phi_{0}\) to provide the closest base instances features set \(\left \{ \phi_{0}\left ( x_{i}^\mathit{base} \right ) \right \}_{i=1}^{k}\) where \(\phi_{0}\left ( x_{i}^\mathit{base} \right )\in \mathbb{R}^{C\times H'\times W'}\). During meta training care is taken so as to exclude the class of the support feature itself from this set of base features so as to force the BaseTransformer to learn to compose novel prototypes using only instances from different classes. These features are reshaped such that the attention would be between spatial locations of \( \phi\left ( x_{i}^{} \right ) \) and spatial locations of the \( \phi_{0}\left ( x_{i}^\mathit{base} \right )\). Key-value pairs of base instance features \(K\phi_{0}\left ( x_{i}^\mathit{base} \right ) \), \(V \phi_{0}\left ( x_{i}^\mathit{base} \right ) \) are obtained using two independent linear layers $K$, $V$ while the transformer's queries \(Q \phi_{}\left ( x_{i} \right ) \) are obtained by using linear mapping $Q$ on the support instance features. Here, we distinguish between a query (or test) set sample and the query of the transformer by explicitly referring to the latter as transformer's query. The dot product between transformer's query and key features results in an attention map between support features and base features. This is followed by a softmax over all spatial locations and $k$ base instances. The computed attention is then used to aggregate the values and a residual connection from the transformer's query features is added to obtain the adapted prototype. Figure \ref{fig:arch_image} illustrates this process.  

We follow the mathematical notation outlined in \cite{doersch2020crosstransformers}. Let \(q_{m}^{i}=Q\phi(x_\mathit{im})\) be the transformer queries i.e., the support features projected by $Q$, where $i$ is the index of the support instance and $m$ is the spatial location and \(k_{n}^{j}=K\phi_{0}(x_{jn}^\mathit{base})\) are the key features, i.e., the base features projected by $K$ where $j$ is the index of the base instance and $n$ is the index of the spatial location. An attention map \(\widetilde{\mathrm{attn}}\) between support features and base features is calculated as: 
\begin{equation}\widetilde{\mathrm{attn}}_{mjn}^{i} = \frac{\exp(\mathrm{attn}^{i}_{mjn})}{\sum_{mjn} \exp(\mathrm{attn}^{i}_{mjn})},\quad
\text{where}\quad \mathrm{attn}_{mjn}^{i} = \langle k^{j}_{n},q^{i}_{m} \rangle.\end{equation}
Next the base adapted prototype \(P_{m}^{i}\) at spatial location $m$ is obtained as follows:
\begin{equation}P_{m}^{i} = q_{m}^{i} + \sum_{jn} \langle\widetilde{\mathrm{attn}}_{mjn}^{i}, v_{n}^{j}\rangle.\end{equation}
For a test instance \(x_{tm}^\mathit{test}\), logits are obtained by calculating the similarity and averaging over the spatial and channel locations as,
\begin{equation}\mathrm{sim}(\phi(x_{t}^\mathit{test}), p^{i}) = - \frac{1}{H'W'}\sum _m \left \| \phi(x_{tm}^\mathit{test}) - P_{m}^{i} \right \|_{2}^{2}.\end{equation}
Here we do not update the features of the base instances during training so as to not corrupt the base data features that have been learnt using several examples per class. The features of a random subset of base instances are computed using the pretrained encoder \(f_{0}\) and stored in a memory bank, which is then queried by the top-$k$ querying function described in Section \ref{sec:queryingfunction}.

\subsection{Querying function} \label{sec:queryingfunction}
We use a semantic similarity based querying function, which uses the label name of the support instance and finds the 5 closest base classes in a semantic space that varies according the dataset. Then base instances are sampled randomly from these classes such that they sum up to $k$. For mini-Imagenet dataset the semantic similarity is equal to the LCH-similarity\cite{leacock1998combining} of the labels in the WordNet graph\cite{miller1995wordnet}. LCH similarity between class labels do not work well for tiered-ImageNet because the class splits were made using higher up nodes in the WordNet hierarchy resulting in very similar LCH similarity scores between a test class label and many base class labels. Hence, we use BERT\cite{devlin-etal-2019-bert} embeddings of the word labels concatenated with their hypernyms from WordNet to find more semantically similar base classes. For CUB, category-level attributes describing the visual features of each bird species are already available. Similar to \cite{shi2019relational}, 
we use the cosine similarity between normalized category attribute vectors to query the closest base classes.

\subsection{Training} \label{sec:training}
Following \cite{doersch2020crosstransformers,mangla2020charting,liu2021learning} we note that we require base embeddings that contain more information than just information regarding base classes to be effective for adapting novel classes. To restrict supervision collapse, we train our encoder with an auxiliary contrastive loss in the pretraining stage. We follow a version of InfoNCE loss from \cite{chen2018a}, where the distance measure is Euclidean instead of cosine distance. 
\begin{equation}l(i,j) = -\log\left(\frac{\exp(s_{i,j})}{\sum_{k=1}^{2N}{1}_{k!=i}\exp(s_{i,k})))}\right),\end{equation}
\begin{equation}L_\text{infoNCE} = \frac{1}{2N}\sum_{k=1}^{N}\left [ l(2k-1,2k) + l(2k,2k-1) \right ],\end{equation}
where \(s_{i,j}=-\left \| f_{i}-f_{j} \right \|_{2}^{2}\) and \(f_{i}\), \(f_{j}\) are features of SimCLR \cite{chen2020simple} style augmented  images in a minibatch.
Concretely, the pretraining is a \(N_{b}\) way classification task where \(N_{b}\) is the number of classes in the base dataset. It is evaluated on a 16-way 1-shot classification task on the validation set. The complete pretraining objective is:
\begin{equation}L_\text{pretraining}=L_\text{classification} + b\times L_\text{InfoNCE},\end{equation}
where $b$ is a hyperparameter balancing the auxiliary loss and \(L_\text{classification}\) is a \(N_{b}\) way cross-entropy loss.


After pretraining, we train the transformer and the encoder end to end in a meta-learning fashion similar to \cite{ye2020few}. Because the feature encoder is pretrained on base dataset, a lower learning rate(factor of 10) is used for the feature encoder to ensure convergence. Similar to the pretraining stage we  use unsupervised InfoNCE loss as an auxiliary loss along with the cross entropy loss during meta training stage to restrict supervision collapse.

\label{sec:blind}

\section{Experiments}

We evaluate our method on three different datasets, namely mini-Imagenet, tiered-Imagenet and CUB \cite{welinder2010caltech}. Mini-Imagenet and tiered-Imagenet are subsets of the Imagenet dataset designed specifically for few shot learning. Mini-Imagenet dataset consists of 60,000 images across 100 classes of which train, validation, and test have 64, 16, and 20 classes respectively. We follow the split specified in \cite{ravi2017optimization}  with 64 classes in the base dataset. Tiered-Imagenet is a larger dataset consisting of 351, 97, and 160 categories for model training, validation, and evaluation, respectively. We follow the split specified in \cite{ye2020few}. In addition to this, we also look at a more fine grained few shot classification task using the CUB dataset that consists of images of various species of birds. CUB dataset contains 11,788 images split into 100, 50, and 50  classes for train, validation, and test. For all images in CUB dataset, we use the provided bounding box to crop all the images as a preprocessing step \cite{triantafillou2017few}. We follow the split specified in \cite{ye2020few}.
 Similar to \cite{ye2020few, rusu2018metalearning}, we use 10,000 randomly sampled few shot tasks for testing as well as report the average accuracy and \(95\%\) confidence intervals.

\section{Implementation details}

We test our method with two networks popularly used in the few shot learning literature, namely Conv4-64 -- a 4 layer convolution network \cite{vinyals2016matching,snell2017prototypical,triantafillou2017few,ye2020few} and ResNet-12 -- a 12-layer residual network \cite{lee2019meta,ye2020few}. As mentioned above we have an additional pretraining stage over the base dataset before the meta training stage. We use images resized to input resolution of $84\times 84$ for both networks. 

In pretraining stage, we use SGD with momentum with an initial learning rate of 0.1 which is decayed by 0.1 using a custom schedule for both networks, similar to \cite{ye2020few}. For weighing the auxiliary contrastive loss, we use balance $b=0.1$. 

In the meta learning stage, we use SGD with momentum with an initial learning rate of 0.002 and $\gamma=20$ for Conv4-64 and an initial learning rate of 0.0002 and $\gamma=40$ for ResNet-12. We follow the standard implementation of multi-headed self attention as presented in \cite{vaswani2017attention}. In meta training stage, the temperature hyperparameter used for softening the logits is critical for convergence to a good solution. We set the temperature as 0.1 for both networks. The optimal value for k is set to 30 after a hyperparameter search.

The memory bank consists of features of 200 randomly sampled instances per base class computed using the trained encoder \(f_0\). The value of $k$ was fixed to be 20 after trying out values of $k$ from 2 to 30 and choosing the best performing value on 1-shot classification on mini-ImageNet.

\begin{table}[]
\centering

\caption{5-way 1-shot and 5-way 5-shot classification accuracy (\%) on miniImageNet dataset using ResNet-12 and Conv4-64 backbones. 95\% confidence intervals reported. The numbers in bold are the best performing methods for the corresponding setting.}
\label{table:mini}
\resizebox{0.8\linewidth}{!}{
\begin{tabular}{lp{1.9cm}p{1.9cm}p{1.9cm}l}

\toprule
Setups          & \multicolumn{2}{c}{1-shot} & \multicolumn{2}{c}{5-shot} \\ 
Backbone        & Conv4-64         & Res12          & Conv4-64         & Res12          \\
\midrule
ProtoNets\cite{snell2017prototypical}       & 49.42$\pm{0.78}$          & 60.37$\pm{0.83}$          & 68.20$\pm{0.66}$           & 78.02$\pm{0.57}$          \\
SimpleShot\cite{wang2019simpleshot}      & 49.69$\pm{0.19}$            & 62.85$\pm{0.20}$           & 66.92$\pm{0.17}$            & 80.02$\pm{0.14}$           \\ 
CAN\cite{hou2019cross}             & -               & 63.85$\pm{0.48}$           & -               & 79.44$\pm{0.34}$         \\    
FEAT\cite{ye2020few}            & 55.15$\pm{0.20}$            & 66.78$\pm{0.20}$           & 71.61$\pm{0.16}$            & 82.05$\pm{0.14}$           \\
DeepEMD\cite{zhang2020deepemd}         & -               & 65.91$\pm{0.82}$           &                 & 82.41$\pm{0.56}$           \\
IEPT\cite{zhang2020iept}         & 56.26$\pm{0.45}$            & 67.05$\pm{0.44}$           & \textbf{73.91$\pm{0.34}$ }  & 82.90$\pm{0.30}$           \\
MELR\cite{fei2020melr}            & 55.35$\pm{0.43}$            & 67.40$\pm{0.43}$           & 72.27$\pm{0.35}$            & 83.40$\pm{0.28}$           \\
InfoPatch\cite{liu2021learning}       & -               & 67.67$\pm{0.45}$           & -               & 82.44$\pm{0.31}$           \\
DMF\cite{xu2021learning}             & -               & 67.76$\pm{0.46}$           & -               & 82.71$\pm{0.31}$           \\
META-QDA\cite{zhang2021shallow}        & 56.41$\pm{0.80}$            & 65.12$\pm{0.66}$           & 72.64$\pm{0.62}$            & 80.98$\pm{0.75}$           \\
PAL\cite{ma2021partner}             & -               & 69.37$\pm{0.64}$          & -               & \textbf{84.40$\pm{0.44}$ }          \\
\midrule
BaseTransformer~~ & \textbf{59.37$\pm{0.19}$ }  & \textbf{70.88$\pm{0.17}$ } & 73.40$\pm{0.18}$            & 82.37$\pm{0.19}$  \\ \bottomrule        
\end{tabular}
}
\end{table}

\subsection{Results}

We report the results of BaseTransformer and other methods for mini-ImageNet in Table \ref{table:mini} and tiered-ImageNet and CUB in Table \ref{minipage:tiered} and \ref{minipage:cub} respectively. We can see that one shot performance of BaseTransformers is better than all competing methods. For fairness, we have excluded comparisons with works that use larger encoders or extra image data \cite{yang2021free}.
We make the following observations. 1. BaseTransformers are effective in improving 1 shot performance on all considered backbones and benchmarks. 2. In comparison to other works \cite{ye2020few,hou2019cross} that use transformers for prototype adaptation we show improvements of 4.1\%, 1.66\%,  and 3.28\% on mini-ImageNet, tiered-ImageNet and CUB dataset in the 1-shot setting. 3. We do not see the strong improvements in 1-shot, reflected in the 5-shot setting. We hypothesize that this could be because the prototypes in 5-shot setting are already a good estimate of the true prototype. We investigate this phenomenon in \ref{sec:5shot}. Results with the oracle top-$k$ querying function are reported in \ref{sec:oracle_q}. See supplementary for comparison with other baselines that use semantic knowledge and detailed results with 95\% confidence intervals.

\begin{table}[]

\begin{minipage}{.4\linewidth}

\caption{5-way 1-shot and 5-way 5-shot classification accuracy (\%) on tieredImageNet dataset for ResNet-12.  The numbers in bold are the best performing methods for the corresponding setting.}
\label{minipage:tiered}
\raggedright
\centering
\resizebox{0.94\linewidth}{!}{
\begin{tabular}{lll}
\toprule

Setups          & 1-shot & 5-shot   \\ 
\midrule
ProtoNets\cite{snell2017prototypical}       & 65.65        & 83.40          \\
SimpleShot\cite{wang2019simpleshot}      & 69.75        & 85.31          \\ 
FEAT\cite{ye2020few}            & 70.80        & 84.79          \\
CAN\cite{hou2019cross}             & 69.89        & 84.23          \\
DeepEMD\cite{zhang2020deepemd}         & 71.16        & 86.03          \\
IEPT\cite{zhang2020iept}         & 72.24        & 86.73          \\
MELR\cite{fei2020melr}            & 72.14        & \textbf{87.01}          \\
InfoPatch\cite{liu2021learning}       & 71.51        & 85.44          \\
DMF\cite{xu2021learning}             & 71.89        & 85.96          \\
META-QDA\cite{zhang2021shallow}        & 69.97        & 85.51          \\
PAL\cite{ma2021partner}             & 72.25        & 86.95          \\
\midrule
BaseTransformer & \textbf{72.46}        &    84.96\\
\bottomrule
\end{tabular}
}
\end{minipage}\hspace{0.4cm}\begin{minipage}{.6\linewidth}

\begin{minipage}{1\linewidth}

\caption{Test accuracy over number of shots for BaseTransformer and SupportTransformer}
\label{bt_st}
\centering
\resizebox{1\linewidth}{!}{%
\begin{tabular}{p{0.15\textwidth}p{0.15\textwidth}p{0.15\textwidth}p{0.15\textwidth}p{0.15\textwidth}l}

\toprule
shot                & 1     & 2     & 3     & 4     & 5     \\
\midrule
BT    & 70.8  & 74.61 & 78.1  & 80.23 & 82.37 \\
ST & 66.34 & 73.12 & 77.33 & 79.8  & 82.01\\
\bottomrule
\end{tabular}
}
\end{minipage}

\vspace{0.5cm}

\caption{5-way 1-shot and 5 way 5-shot classification accuracy (\%) on CUB dataset. The numbers in bold are the best performing methods for the corresponding setting.}
\label{minipage:cub}
\centering
\resizebox{\linewidth}{!}{%
\begin{tabular}{lllll}
\toprule
Setups          & \multicolumn{2}{l}{1-shot} & \multicolumn{2}{l}{5-shot} \\ 
Backbone        & Conv4-64         & Res12          & Conv4-64          & Res12         \\
\midrule 
ProtoNets\cite{snell2017prototypical}       & 64.42           & -              & 81.82            & -             \\
FEAT\cite{ye2020few}            & 68.87           & -              & 82.90            & -             \\
DeepEMD\cite{zhang2020deepemd}         & -               & 75.65          & -                & 88.69         \\
IEPT\cite{zhang2020iept}         & 69.97  & -              & 84.33            & -             \\
MELR\cite{fei2020melr}            & 70.26           & -              & \textbf{85.01}            & -             \\
\midrule 
BaseTransformer & \textbf{72.15}  & \textbf{82.27} & 82.12             & \textbf{90.64}  \\ 
\bottomrule
\end{tabular}%
}

\end{minipage}

\end{table}

\subsection{Ablation studies}
Table \ref{table:ablation} provides detailed ablation study of the various parts of our method for the Conv4-64 encoder. We can see that performance without BaseTransformer and SimCLR-pretraining is similar to that of Prototypical Networks. Including just InfoNCE as the auxiliary loss in the pretraining stage improves performance by 1.3\%. Applying BaseTransformers with visual querying on Prototypical networks further improves one shot accuracy to 54.46\%. Using SimCLR in the pretraining stage with BaseTransformers improves accuracy further to 57.38\%. This shows that the SimCLR loss in the pretraining stage is necessary to prevent supervision collapse and provide the BaseTransformer with robust base features. Finally, applying semantic querying gives a further improvement of $\sim 2\%$.

\subsection{5 shot results}\label{sec:5shot}
We believe that the performance improvements from using base dataset is only significant in the 1 shot to 3 shot domain. We ran experiments comparing BaseTransformer (BT) with semantic querying to SupportTransformer(ST)- a variant of BT where the \(Q=\sum_{y_{i}\in c}^{} \phi\left ( x_{i} \right )\) and \(K = V =\left \{ \phi(x_{i}) \right \}\)  where \(y_{i} \in c\), keeping all other hyperparameters same. Here Q = prototype of class c and K=V= support instances of class c.
Test accuracy of ST approaches that of BT as the number of shots approach 5, showing that the prototypes from 5 different support instances of the novel class become as good as the prototype computed using base instances queried via semantic query as seen in Table \ref{bt_st}.

\subsection{Oracle querying}\label{sec:oracle_q}
Table \ref{table:oracle} reports 1-shot classification results using visual, semantic, and oracle querying for the mini-ImageNet and tiered-ImageNet datasets. Oracle querying uses the ResNet-12 encoder trained on both seen and unseen classes in the dataset. Then the closest base classes are found by the Euclidian similarity between the class prototypes estimated using all the instances in the class. We see that by improving the querying function, BaseTransformers can improve 1-shot accuracy by a significant margin, especially for the tiered-ImageNet dataset where the classes are distributed into seen and unseen classes with limited semantic overlap \cite{oreshkin2018tadam}. This shows that the 1-shot performance of the BaseTransformer architecture is limited by the querying function. We leave the search for an optimal querying function for the future.

\begin{wrapfigure}{L}{0.5\textwidth}
\includegraphics[width=6.3cm]{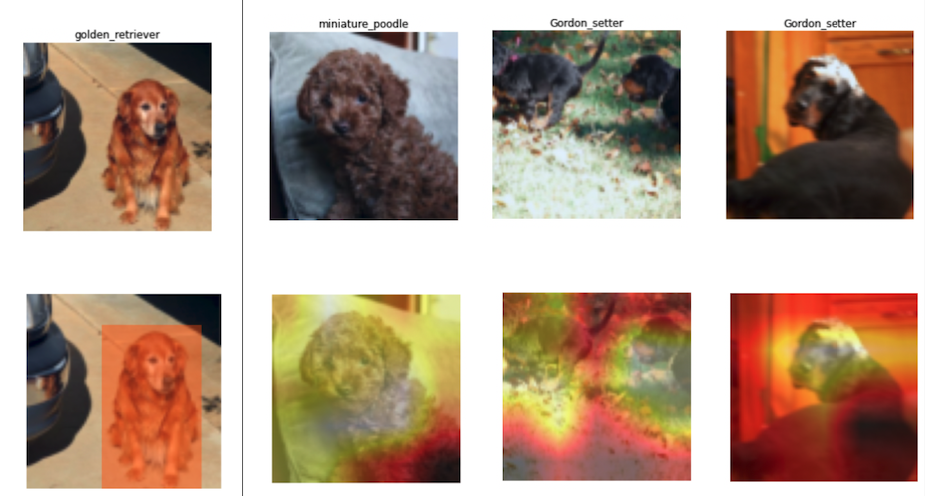}
\caption{Left: support instance; right: the three closest base instances (top) and attention maps overlaid over the closest base instances (bottom).}
\label{fig:attn-maps}

\end{wrapfigure}

\subsection{Visualization of learnt attention over base datapoints}
 We visualize the attention maps learnt by the BaseTransformer in Fig.~\ref{fig:attn-maps}. These are obtained by overlaying the resized attention map over the corresponding image of base instance selected by the querying function.  We can see that for each support image, BaseTransformer has learnt to attend to semantically similar regions of base instances. BaseTransformer is successful in identifying multiple visually similar features in base instance images when there are multiple instances of the class in one image. For example for golden retriever, the BaseTransformer attends to two instances of gordon setter without being explicitly trained to identify multiple gordon setters -- see supplementary material for more examples.

\begin{table}[]
\begin{minipage}{0.5\linewidth}
\caption{1 shot results using oracle querying function}
\label{table:oracle}
\resizebox{\textwidth}{!}{%
\begin{tabular}{lll}
\toprule
Querying Setup & mini-ImageNet & tiered-ImageNet \\
\midrule
Visual         & 67.40$\pm${0.20}         &  71.05$\pm${0.18}\\
Semantic       & 70.88$\pm${0.17}         & 72.46$\pm${0.19}           \\
Oracle         & 72.38$\pm${0.18}         & 76.55$\pm${0.17}      \\
\bottomrule
\end{tabular}%
}

\end{minipage}\hspace{0.01\linewidth}\begin{minipage}{0.5\linewidth}

\caption{Ablation of various components of BaseTransformer}
\label{table:ablation}
\centering
\resizebox{0.7\linewidth}{!}{
\begin{tabular}{llll}
\toprule
SimCLR-pre & Querying & BT & 1-shot \\
\midrule
No         & NA             & No              & 51.65  \\
Yes        & NA             & No              & 52.68  \\
No         & Visual             & Yes             & 54.46       \\
Yes        & Visual             & Yes             & 57.38  \\
Yes        & Semantic            & Yes             & 59.37 \\
\bottomrule
\end{tabular}
}
\end{minipage}
\end{table}

\section{Conclusion}

In this paper we propose that the one shot performance of metric learning based few shot approaches is hindered by the bias in estimation of the prototype. We show that prototype estimation can be improved by using well supported base instance features. Our proposed method, BaseTransformers, adapts the prototype by making use of learnt correspondences between the support instance and closest base class instances. Extensive experiments on three benchmarks and two encoders show the effectiveness of our method.

\section*{Acknowledgement}
This publication has emanated from research conducted with the financial
support of Science Foundation Ireland under Grant number 18/CRT/6183. For the purpose
of Open Access, the author has applied a CC BY public copyright licence to any
Author Accepted Manuscript version arising from this submission.

\bibliography{egbib}
\end{document}


\maketitle

\section{5-shot experiments}

For the 5-shot case, we experiment with two different ways of averaging the support instances to form a prototype. Pre-avg averages the support instances before the BaseTransformer. The closest base instances in this case are sampled randomly from the 5 closest base classes using semantic similarity as described in Section 3.4 in the main paper. In contrast, for post-avg we adapt each support instance and its corresponding set of closest base instances  independently and the prototype is obtained by averaging the adapted support instances after the BaseTransformer. Table \ref{table:5shot} reports the results for both pre-avg and post-avg for 5-shot classification on the mini-ImageNet dataset using a ResNet12 encoder. Here we can see that pre-avg  works much better than post-avg for the ResNet12 encoder. We believe that this could be because averaging the support instances results in a more robust input to the BaseTransformer, aiding in its training.  
\begin{table}[]

\begin{minipage}{0.4\linewidth}
\caption{Results for different setups considered for averaging of support instances in 5-shot setting.}
\label{table:5shot}
\centering
\begin{tabular}{p{3cm}l}
\toprule
Setup    & 5-shot \\
\midrule
Pre-avg  & 78.38$\pm${0.23}  \\
Post-avg & 82.05$\pm${0.19}  \\
\bottomrule   
\end{tabular}
\end{minipage}\qquad \begin{minipage}{0.55\linewidth}
\caption{Comparison with semantic knowledge baselines}
\label{table:semantic}
\resizebox{\textwidth}{!}{%
\begin{tabular}{llll}
Method          & mini 1-shot & CUB 1-shot & tiered 1-shot \\
\midrule
RS\_FSL \cite{afham2021rich}         & 65.33$\pm${0.83} & 65.66$\pm${0.90} & -             \\
MS \cite{schwartz2022baby} & 67.3       & 76.1       & -             \\
AM3 \cite{xing2019adaptive}             & 65.30$\pm${0.49} & 74.1       & 69.08$\pm${0.47}    \\
KTN \cite{peng2019few}             & 64.42      & -          & -             \\
\midrule
BT (Ours)        & 70.88$\pm${0.17} & 82.27$\pm${0.19} & 72.46$\pm${0.19}   
\bottomrule
\end{tabular}\vspace{-0.5em}
}
\end{minipage}

\end{table}

\begin{table}[t]

\caption{5-way 1-shot and 5-way 5-shot classification accuracy (\%) on tiered-ImageNet dataset for ResNet-12.  The numbers in bold are the best performing methods for the corresponding setting.}

\label{table:tiered}
\raggedright
\centering
\begin{tabular}{lll}
\toprule

Setups          & 1-shot & 5-shot   \\ 
\midrule
ProtoNets \cite{snell2017prototypical}       & 65.65$\pm${0.92}        & 83.40$\pm${0.65}          \\
SimpleShot \cite{wang2019simpleshot}      & 69.75$\pm${0.20}        & 85.31$\pm${0.15}          \\ 
FEAT \cite{ye2020few}            & 70.80$\pm${0.23}        & 84.79$\pm${0.16}          \\
CAN \cite{hou2019cross}             & 69.89$\pm${0.51}        & 84.23$\pm${0.37}          \\
DeepEMD \cite{zhang2020deepemd}         & 71.16$\pm${0.87}        & 86.03{0.58}          \\
IEPT \cite{zhang2020iept}         & 72.24$\pm${0.50}        & 86.73$\pm${0.34}          \\
MELR \cite{fei2020melr}            & 72.14$\pm${0.51}        & \textbf{87.01}$\pm${0.35}          \\
InfoPatch \cite{liu2021learning}       & 71.51$\pm${0.52}        & 85.44$\pm${0.35}          \\
DMF \cite{xu2021learning}             & 71.89$\pm${0.52}        & 85.96$\pm${0.35}          \\
META-QDA \cite{zhang2021shallow}        & 69.97$\pm${0.52}        & 85.51$\pm${0.58}          \\
PAL \cite{ma2021partner}             & 72.25$\pm${0.72}        & 86.95$\pm${0.47}          \\
\midrule
BaseTransformer & \textbf{72.46}$\pm${0.19}        &    84.96$\pm${0.18}\\
\bottomrule
\end{tabular}
\end{table}

\begin{table*}[t]

\caption{5-way 1-shot and 5 way 5-shot classification accuracy (\%) on CUB dataset. The numbers in bold are the best performing methods for the corresponding setting.}
\label{table:cub}
\centering
\begin{tabular}{lllll}
\toprule
Setups          & \multicolumn{2}{l}{1-shot} & \multicolumn{2}{l}{5-shot} \\ 
Backbone        & Conv4-64         & ResNet12          & Conv4-64          & ResNet12         \\
\midrule 
ProtoNets \cite{snell2017prototypical}       & 51.31$\pm${0.89}           & 66.09$\pm${0.92}              & 70.77$\pm${0.70}           & 82.50$\pm${0.58}             \\
FEAT \cite{ye2020few}            & 68.87$\pm${0.22}           & -              & 82.90$\pm${0.15}            & -             \\
DeepEMD \cite{zhang2020deepemd}         & -               & 75.65$\pm${0.83}          & -                & 88.69$\pm${0.50}         \\
IEPT \cite{zhang2020iept}         & 69.97$\pm${0.49}  & -              & 84.33$\pm${0.33}            & -             \\
MELR \cite{fei2020melr}            & 70.26$\pm${0.50}           & -              & \textbf{85.01}$\pm${0.32}            & -             \\
\midrule 
BaseTransformer & \textbf{72.15}$\pm${0.20}  & \textbf{82.27}$\pm${0.19} & 82.12$\pm${0.21}             & \textbf{90.64}$\pm${0.18}  \\ 
\bottomrule
\end{tabular}%
\end{table*}

\section{Tiered-ImageNet and CUB detailed results}
Detailed results with 95\% confidence intervals are reported for the tiered-ImageNet and CUB datasets in Table \ref{table:tiered} and Table \ref{table:cub} respectively. We have reported tiered-ImageNet results only for the ResNet12 encoder. 

\section{Comparison with Semantic knowledge baselines}

Previous methods that use semantic knowledge \cite{schwartz2022baby,afham2021rich,xing2019adaptive,peng2019few} use it explicitly to structure the feature space, while we use it only for querying. Despite this, our method outperforms all these methods (see \ref{table:semantic}).

\section{Visualization of learnt attention over base datapoints}

\begin{figure*}[!t]
\centering
\includegraphics[width=12cm]{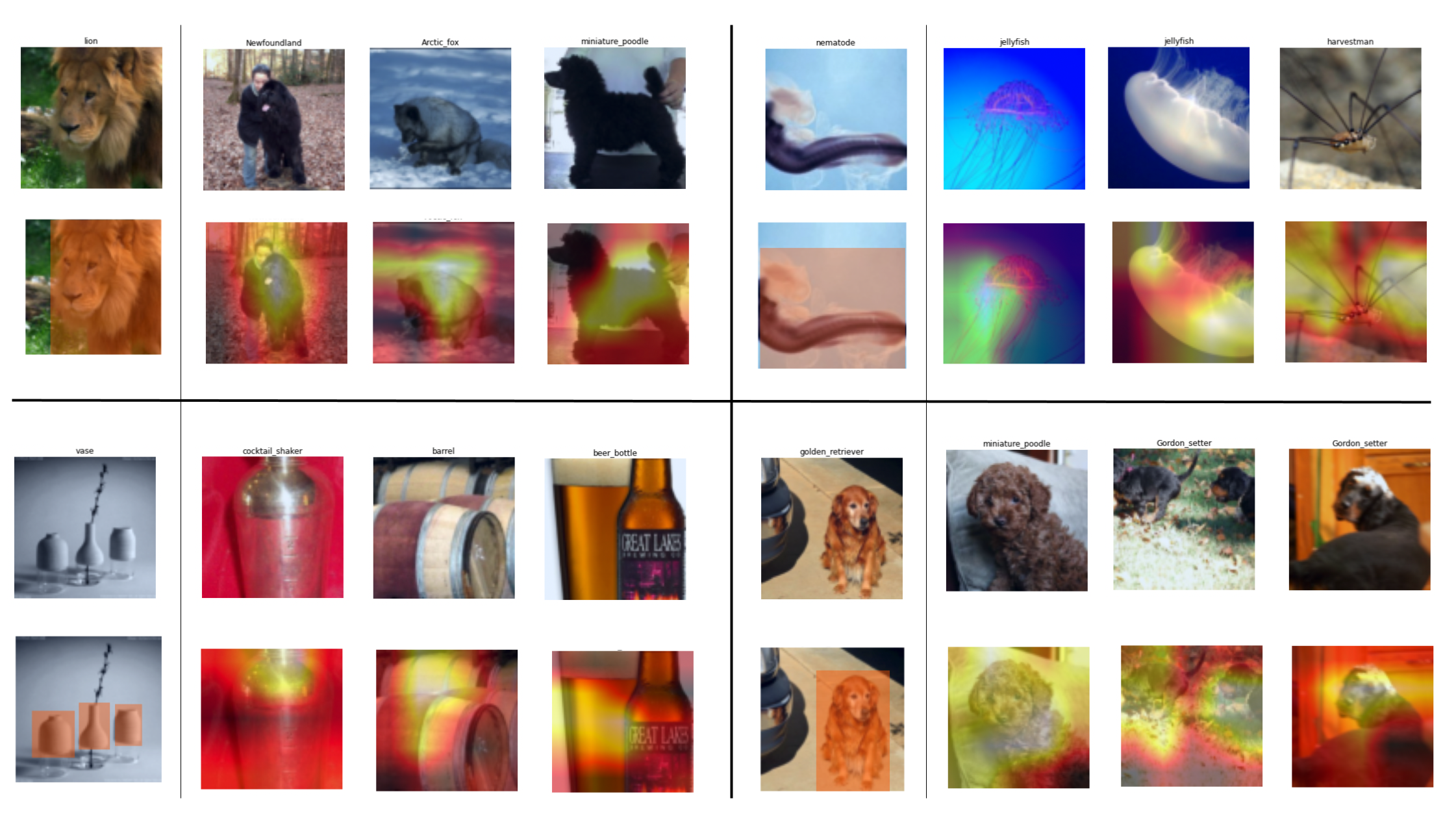}
\caption{ Left: support instance; right: the three closest base instances (top) and attention maps overlaid over the closest base instances (bottom). It can be seen that BaseTransformers learns to select visually similar features from the base feature space using the learnt part based correspondences. Warmer color corresponds to higher attention weight. }
\label{fig:attn-maps}
\end{figure*}

We visualize the attention maps learnt by the BaseTransformer in Fig.~\ref{fig:attn-maps}. These are obtained by overlaying the resized attention map over the corresponding image of base instance selected by the querying function.  We can see that for each support image, BaseTransformer has learnt to attend to visually similar regions of base instances. For example (Fig.~1 quadrant 2), for support instance nematode, the BaseTransformer learns to attend to the tentacle of jelly fish or the legs of harvestman to improve the prototype representation. It is also worth noting that in some cases the BaseTransformer is successful in identifying multiple visually similar features in base instance images when there are multiple instances of the class in one image. For example (Fig.~1 quadrant 4), for golden retriever, the BaseTransformer attends to two instances of gordon setter without being explicitly trained to identify multiple gordon setters.

\bibliography{egbib}